\documentclass{article}

    \usepackage[nonatbib,preprint]{neurips_2021}

\usepackage[utf8]{inputenc} 
\usepackage[T1]{fontenc}    
\usepackage{hyperref}       
\usepackage{url}            
\usepackage{booktabs}       
\usepackage{amsfonts}       
\usepackage{nicefrac}       
\usepackage{microtype}      
\usepackage{multicol,lipsum}
\usepackage{subcaption}
\usepackage[skip=2pt,font=footnotesize]{caption}
\usepackage{comment}
\usepackage{booktabs, tabularx} 
\usepackage[numbers]{natbib}
\usepackage{times}
\usepackage{url}
\usepackage{algorithm}
\usepackage{algorithmic}
\usepackage{graphicx}
\usepackage{mathtools}
\usepackage{amsmath}
\usepackage{amssymb}
\usepackage{amsthm} 
\usepackage{thmtools}
\usepackage{hyperref}
\usepackage[capitalise,nameinlink,noabbrev]{cleveref}
\usepackage{enumitem}
\usepackage{float}
\usepackage[export]{adjustbox}
\usepackage{xcolor,colortbl}
\usepackage{amsfonts}
\usepackage{bm}
\usepackage{xpatch}
\usepackage{wrapfig}
\usepackage[symbol]{footmisc}
\usepackage{pifont}

\DeclarePairedDelimiterX{\inp}[2]{\langle}{\rangle}{#1, #2}

\newcommand{\ip}[1]{\langle #1 \rangle}
\DeclareMathOperator*{\1}{\mathbbm{1}}

\usepackage{listings}
\usepackage{color}
\usepackage{makecell}

\definecolor{dkgreen}{HTML}{757575}
\definecolor{gray}{HTML}{000000}
\definecolor{mauve}{HTML}{0097A7}

\def\1{\bm{1}}

\DeclareMathAlphabet{\mathsfit}{\encodingdefault}{\sfdefault}{m}{sl}
\SetMathAlphabet{\mathsfit}{bold}{\encodingdefault}{\sfdefault}{bx}{n}

\newcommand{\I}{\mathbb{I}}

\newcommand*\norm[1]{\left\|#1\right\|}

\usepackage{xcolor}

\newtheorem{theorem}{Theorem}
\newtheorem{proposition}{Proposition}
\newtheorem{definition}{Definition}
\newtheorem{assumption}{Assumption}
\newtheorem{lemma}{Lemma}

\title{Model-Invariant State Abstractions for Model-Based Reinforcement Learning}

\author{%
  Manan Tomar \\
  University of Alberta
  \And
   Amy Zhang \\
   FAIR, Menlo Park
   \AND
   Roberto Calandra \\
   FAIR, Menlo Park
   \And
   Matthew E. Taylor \\
   University of Alberta
   \And
   Joelle Pineau \\
   FAIR, Montreal
}

\begin{document}

\maketitle

\begin{abstract}
    Accuracy and generalization of dynamics models is key to the success of model-based reinforcement learning (MBRL). As the complexity of tasks increases, so does the sample inefficiency of learning accurate dynamics models. However, many complex tasks also exhibit sparsity in the dynamics, i.e., actions have only a local effect on the system dynamics. In this paper, we exploit this property with a causal invariance perspective in the single-task setting, introducing a new type of state abstraction called \textit{model-invariance}. 
    Unlike previous forms of state abstractions, a model-invariance state abstraction  leverages causal sparsity over state variables. This allows for compositional generalization to unseen states, something that non-factored forms of state abstractions cannot do.
    We prove that an optimal policy can be learned over this model-invariance state abstraction and show improved generalization in a simple toy domain.
    Next, we propose a practical method to approximately learn a model-invariant representation for complex domains and validate our approach by showing improved modelling performance over standard maximum likelihood approaches on challenging tasks, such as the MuJoCo-based Humanoid.
    Finally, within the MBRL setting we show strong performance gains with respect to sample efficiency across a host of other continuous control tasks. 
\end{abstract}

\section{Introduction}


Model-based reinforcement learning~(MBRL) is a popular framework for data-efficient learning of control policies.
At the core of MBRL is learning an environmental dynamics model and using it to:
1) fully plan~\citep{deisenroth2011pilco, chua2018deep}, 
2) augment the data used by a model-free solver~\citep{sutton1991dyna}, or 
3) use as an auxiliary task while training~\citep{lee2019stochastic, zhang2021dbc}. 
However, learning a dynamics model --- similar to other supervised learning problems --- suffers from the issue of generalization, since the data we train on is not necessarily the data we test on. 
This is a persistent issue that is worsened in MBRL as even a small inaccuracy in the dynamics model or changes in the control policy can result in visiting completely unexplored parts of the state space. 
Thus, it is generally considered beneficial to learn models capable of generalizing well.
Various workarounds for this issue have been explored in the past; for example, coupling the model and policy learning processes \citep{lambert2020objective} so that the model is always accurate to a certain threshold, or using an ensemble of models to handle the uncertainty in each estimate~\citep{chua2018deep}. 
However, these approaches are unnecessarily pessimistic, and do not leverage special structure in dynamics for better generalization.

\begin{wrapfigure}{r}{0.35\linewidth}
\vspace{-5pt}
    \centering
    \includegraphics[width=0.4\linewidth]{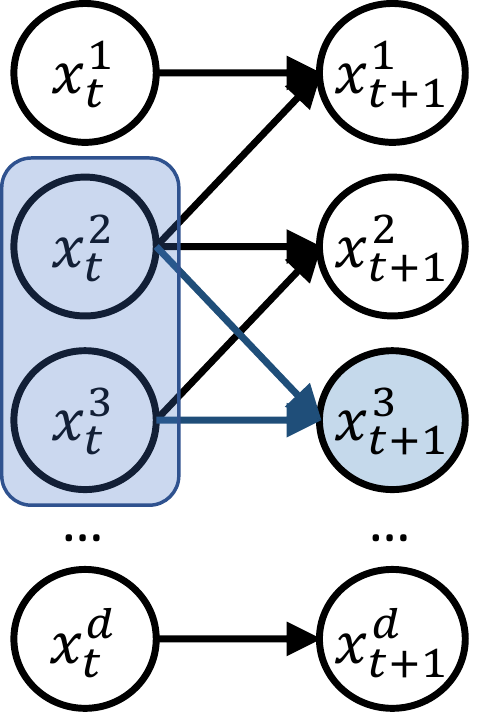}
    \caption{Graphical model of sparsity across state variables. For example, the dimension $x^{3}_{t+1}$ (shaded in blue) only depends on two dimensions $x^{3}_{t}$ and $x^{2}_{t}$ (in the blue box).}
    \label{fig: setup}
\vspace{-5pt}
\end{wrapfigure}

This paper studies how to improve generalization capabilities through state abstraction.
In particular, we leverage two existing concepts to define a new type of state abstraction that yields improved generalization performance for MBRL. 
The first concept is that many real world problems exhibit \textit{sparsity} in the local dynamics --- given a set of state variables, each variable only depends on a small subset of those variables in the previous timestep (see Figure~\ref{fig: setup}).
The second concept is the principle of causal invariance, which dictates that given a set of features, we should aim to build representations that comprise \textit{only} those features that are consistently necessary for predicting the target variable of interest across different interventions~\citep{peters2015causal}. In the MBRL context, we can cast model learning with causal invariance as an optimization objective where the target variables are the next state variables and input features are the current state and action variables (the probable set of causal predictors of the target).
The intuition is that since we learn a predictor that only takes into account invariant features that consistently predict the target variable well, it is likely to contain the true causal features, and therefore, will generalize well to all possible shifts in the data distribution. 
The two concepts of sparsity and causality are intertwined in that they both are forms of inductive biases that surround the agent dynamics~\citep{goyal2020inductive}. The policy of a continuously improving learner plays an important role, as it allows us to realize both causal invariance and sparsity. 
In this context, we ask the question, \emph{can we exploit sparsity in the transition dynamics to learn a model with improved generalization ability to unseen parts of the state-action space?}
Ultimately, based on experimental results, and using causal invariance as a tool, we will show that the answer is ``\textit{yes}.''
Given basic exploratory assumptions, we show both theoretically and empirically that we can learn a model that generalizes well on state distributions induced by any policy distinct from the ones used while learning it.

The contributions of this paper are as follows. We, {\bf 1)}
highlight an important condition required to answer the above raised question, that of independence between state variables in a dynamics model. This observation leads to the proposal of a new kind of state abstraction, \textit{model-invariance}. Model-invariance leverages natural sparsity over state variables by constructing coarser state abstractions on a per-variable level, allowing for new generalization capabilities over novel compositions of state variable values; {\bf 2)} show that a representation that only uses the causal parents of each state variable is, in fact, a model-invariant representation; {\bf 3)} show that learning a model over such an abstraction, and then planning using this model, is optimal, given certain exploratory assumptions; {\bf 4)} perform a proof-of-concept experiment in the batch setting to show that such a model learning approach always leads to better generalization in unseen parts of the state space; {\bf 5)} then introduce a practical method which approximates learning a model-invariant representation for more complex domains which use function approximation; {\bf 6)} empirically show that our approach results in better model generalization for domains such as the MuJoCo-based Humanoid and follow this by combining our model learning scheme with a policy optimization framework, which leads to improvements in sample efficiency. We believe that the proposed algorithm is an important step towards leveraging sparsity in complex environments and to improve generalization in MBRL methods.



\section{Preliminaries}
We consider the agent's interaction with the environment as a discrete time $\gamma$-discounted Contextual Decision Process~(CDP), a term recently proposed by \citet{krishnamurthy2016pac} to broadly model sequential decision processes which require the policy to be based on rich features (context). A CDP is defined as $\mathcal{M}=(\mathcal{X}, \mathcal{A},P,R,\gamma,\mu_0)$, where $\mathcal{X} \subseteq \mathbb{R}^{d}$ is a finite state space and $\mathcal{A}$ is the and action space; $P \equiv P(x'|x,a)$ is the transition kernel; $R \equiv r(x,a)$ is the reward function with the maximum value of $R_{\text{max}}$; $\gamma\in(0,1)$ is the discount factor; and $\mu_0$ is the initial state distribution. CDPs generalize MDPs by unifying decision problems that depend on rich context. Let $\pi: \mathcal{X}\rightarrow \Delta_{\mathcal{A}}$ be a stationary memoryless policy, where $\Delta_{\mathcal{A}}$ is the set of probability distributions on $\mathcal{A}$. The discounted frequency of visiting a state $s$ by following a policy $\pi$ is defined as $\rho_\pi(x)\equiv (1-\gamma) \mathbb{E}[\sum_{t\geq 0}\gamma^t \mathbb I \{x_t=x\}\mid \mu_0, \pi]$. The value function of a policy $\pi$ at a context $x\in\mathcal{X}$ is defined as $V^\pi(x) \equiv \mathbb{E}[\sum_{t\geq 0}\gamma^tr(x_t,a_t)|x_0=x, \pi]$. Similarly, the action-value function of $\pi$ is defined as $Q^{\pi}(x,a) = \mathbb{E}[\sum_{t\geq 0}\gamma^tr(x_t,a_t)|x_0=x,a_0=a,\pi]$. The Bellman optimality operator $\mathcal{T}: \mathbb{R}^{|\mathcal{X} \times \mathcal{A}|} \rightarrow \mathbb{R}^{|\mathcal{X} \times \mathcal{A}|}$ is defined as $\mathcal{T} Q(x, a) = r(x, a) + \gamma \ \inp[\big]{P(\cdot | x, a)}{\max_{a'} Q(\cdot, a')}$. The CDP definition also assumes that there exists a set of latent states~$\mathcal{S}$. The notion of $S$ becomes more apparent if further structural assumptions are posed, such as that of a Block MDP~\citep{du2019provably, zhang2020invariant}. 

\subsection{Setup}
\label{subsec: problem-formulation}
There are two important cases we can consider with CDPs. We explore these with simple examples:

\textbf{Case 1}: Large state space or full state input: Consider $\mathcal{X}$ as the proprioceptive states of a robot. $\mathcal{X}$ is not a rich observation, but rather an arbitrarily large  set of state variables $\{x^1,x^2,...,x^d\}$. There is likely to be little to no irrelevant information present w.r.t. the downstream task in such a case, and hence the latent state space and observation space are of the same dimension, i.e., $\mathcal{S}, \mathcal{X} \subseteq \mathbb{R}^{d}$, but $\mathcal{S}$ could potentially contain a lot fewer number of states than $\mathcal{X}$, i.e., $|\mathcal{S}| \ll |\mathcal{X}|$. Here, $\mathcal{S}$ corresponds to some coarser abstraction of the given full state, learning and planning over which can still be optimal. 

\textbf{Case 2}: Rich observation or pixel based input: Consider $\mathcal{X}$ to be a set of images, for example, each being a frontal camera view of a robot. There is irrelevant information present in the form of background pixels. Nevertheless, the latent state set $\mathcal{S}$ is still the same as in the previous case, a coarse representation of the rich observation space $\mathcal{X}$. This task is more challenging, in that we first have to encode a low-dimensional state of the robot from the image that exhibits sparsity (equivalent to what is handed to us a priori in case 1 in the form of $\mathcal{X}$) and then learn a model-invariant representation $\mathcal{S}$. 

In this work, we focus on Case~1 and from now on use the term CDP and MDP interchangeably throughout the paper. 
However, we remain general in our setup description since Case~2 becomes immediately relevant if we have a way to learn a compressed representation with sparseness properties, which makes our method applicable. 
In both cases, we assume that the transition dynamics over the full state are factorized. More formally:
\begin{assumption} (Transition Factorization)
\label{asp: transition factorization}
For given full state vectors $x_t, x_{t+1} \in \mathcal{X}$, action $a \in \mathcal{A}$, and $x^i$ denoting the $i^{\text{th}}$ dimension of state $x$ we have $P(x_{t+1} | x_t, a) = \prod_{i} P(x^{i}_{t+1} | x_t, a)$.
\end{assumption}

Note that this is a weaker assumption than factored MDPs \citep{kearns1999efficient, guestrin2001max} as we do not assume a corresponding factorization of the reward function. Also note that in either case, the factorization property is only assumed on the full state input and not on the rich observation.



\subsection{Invariant Causal Prediction}
\label{subsec: invariant-causal-prediction}

Invariant causal prediction (ICP) \citep{peters2015causal} considers learning an invariant representation w.r.t. spurious correlations (see Appendix~\ref{sec: Motivation}) that arise due to noise in the underlying (unknown) causal model describing a given system. The key observation is that if one considers the direct causal parents of a response/target variable of interest ($Y$), then the conditional distribution of $Y$ given these direct causes $\textbf{PA}(Y)$ does not change across interventions on any variable except $Y$. Therefore, ICP suggests collecting data into different environments (corresponding to different interventions), and output the set of variables $X_i$ for which a learned predictor of $Y$ remains the same given $X_i$ across the multiple environments with high probability. 

\subsection{State Abstractions and Model Irrelevance}
\label{subsec: state-abstractions}

State abstractions allow us to map behaviorally-equivalent states into a single abstract state, thus simplifying the learning problem, which then makes use of the (potentially much smaller set of) abstract states instead of the original states~\citep{bertsekas1988adaptive}. In principle, any function approximation architecture can act as an abstraction, since it attempts to group similar states together. Therefore, it is worth exploring the properties of a representation learning scheme as a state abstraction. In the next section, we build our theory based on this connection. 

A specific kind of state abstraction interesting to us is the model irrelevance state abstraction or bisimulation \citep{even2003approximate, ravindran2004algebraic, li2009unifying}. An abstraction $\phi:\mathcal{X}\mapsto\mathcal{S}$ is model irrelevant if for any two states $x_1,x_2$ and next state $x \in \mathcal{X}$, abstract state $s \in \mathcal{S}$, $a \in \mathcal{A}$ where $\phi(x_1) = \phi(x_2)$, we have $R(x_1, a) = R(x_2, a)$, and $\sum_{x \in \phi^{-1}(s)} P(x | x_1, a) = \sum_{x \in \phi^{-1}(s)} P(x | x_2, a)\,.$
%
%
%
Since an exact equivalence is not practical, prior work deals with approximate variants through the notion of $\epsilon$-closeness~\citep{jiang2018notes}.
The main difference between a model irrelevance state abstraction and our proposed model-invariance state abstraction is that the model irrelevance abstraction does not leverage sparsity in factored dynamics. Our model-invariance state abstraction is variable specific, assuming the state space consists of a set of state variables. We formally define our model-invariance state abstraction in the following section.

\section{Model Invariance and Abstractions}
\label{subsec: inv_intro} 

In this section, we build towards our goal of learning a generalizable transition model, given limited environmental data. However, we first highlight how the independence assumption (Assumption~\ref{asp: transition factorization}) connects to this central goal by introducing a new kind of state abstraction called model-invariance.

Given conditional independence over state variables, we define model-invariance as an abstraction that preserves transition behavior for each state variable. Formally, we define a reward-free version as follows:



\begin{definition} (Model Invariance)
\label{def: model-invariant-abstraction}
$\phi_i$ is model-invariant if $\forall x, x_1,x_2 \in \mathcal{X},  a \in \mathcal{A}$, $\phi_{i}(x_1) = \phi_{i}(x_2)$ if and only if $P(x^i | x_1, a) = P(x^i | x_2, a)$, where $x^i$ denotes the value of state variable $i$ in state $x$. 
\end{definition}
In words, an invariant abstraction $\phi_i$ is one which has the same transition probability to next state for any two given states $x_1$ and $x_2$, in the $i^{\text{th}}$ index.  If we assume factored rewards, we can define a corresponding reward-based invariant abstraction that parallels the model-irrelevance abstraction more closely, but we focus here on the reward-free setting. 

Since it is impractical to ensure this equivalence exactly, we introduce an approximate definition which ensures an $\epsilon$-closeness. 
\begin{definition} (Approximate Model Invariance)
$\phi$ is $\epsilon_{i, P}$-model-invariant if for each index $i$, 
\begin{equation*}
    \sup_{\substack{a\in\mathcal{A}, x,x_1,x_2 \in \mathcal{X},  \phi(x_1):=\phi(x_2)}} \norm{P(x^i | x_1, a) - P(x^i | x_2, a)} \leq \epsilon_{i, P}.
\end{equation*}
$\phi$ is $\epsilon_R$-model-invariant if $\epsilon_R =\sup_{\substack{a\in\mathcal{A},x_1,x_2\in \mathcal{X},\phi(x_1)=\phi(x_2)}} \big|R(x_1,a)-R(x_2,a)\big|\,.$
\end{definition}
%

\begin{lemma} (Model Error Bound)
\label{lemma: model-error-bound}
Let $\phi$ be an $\epsilon_{i,P}$-approximate model-invariant abstraction on CDP $M$. Given any distributions ${p_{s_i} : s_i \in \phi_i(\mathcal{X})}$ where $p_{s_i}$ is supported on $\phi^{-1}({s_i})$ and $p_s = \prod^{d}_{i=1} p_{s_i}$, we define $M_{\phi} = ({\phi(\mathcal{X})}, \mathcal{A}, P_{\phi}, R_{\phi}, \gamma)$ where $P_{\phi}(s, a) = \mathbb{E}_{x \sim p_s} [P(\cdot | x, a)] $ and $R_{\phi} = \mathbb{E}_{x \sim p_s} [R(x, a)]$. Then for any $x \in \mathcal{X}$, $a \in \mathcal{A}$,
\vspace{-0.5cm}
\begin{align*}
    \norm{P_{\phi}(s, a) - \Phi P(x, a)} &\leq {\displaystyle \sum^{d}_{i=1}} \epsilon_{i, P},
\end{align*}
\vspace{-0.5cm}
\end{lemma}
where $\Phi P$ denotes the \textit{lifted} version of $P$, where we take the next-step transition distribution from observation space $\mathcal{X}$ and lift it to latent space $\mathcal{S}$ (Proof in Appendix~\ref{app-sec: proofs}). \cref{lemma: model-error-bound} provides a bound on the modelling error when the individual errors for an approximate model-invariant abstraction are compounded. Specifically, $P_{\phi}$ refers to the transition probability of a CDP that acts on the states $\Phi(\mathcal{X})$, rather than the original CDP that acts on the original states.
Note that we are particularly concerned with the case where each $x_{i}$ is atomic in nature, i.e., it is not further divisible. Such a property ensures that model-invariance does not collapse to model irrelevance. 

\section{Causal Invariance in Model Learning}

We now move on to providing a connection between causal invariance and model-invariant abstractions. First, we describe the causal setup below: 


\begin{definition} (Causal Setup, \citep{peters2015causal})
\label{def: causal-setup}
For each future state variable indexed by $i$, $x^{i}_{t+1}$, there exists a linear structural equation model consisting of state dimensions and actions, $(x^{i}_{t+1}, x^{1}_t, ..., x^{d}_t, a_t)$ with coefficients $(\beta_{jk})_{j,k=1, ...,d+2}$, given by a directed acyclic graph. An experimental setting $e \in \mathcal{E}$ arises due to one or more interventions on the variable set $\{x^{1}_t, ..., x^{d}_t, a_t\}$, with the exception of $x^{i}_{t+1}$. 
\end{definition}

\begin{assumption} (Invariant Prediction)
\label{asp: invariant}
For each $e \in \mathcal{E}$: the experimental setting $e$ arises due to one or several interventions on variables from $(x^{1}_t, ..., x^{d}_t, a_t)$ but not on $x^{i}_{t+1}$; here, we allow for do-interventions~\citep{pearl2009causality} or soft-interventions~\citep{eberhardt2007interventions}. 
\end{assumption}

For our purposes, each intervention corresponds to a change in the action distribution, i.e., policy. Thus, in turn, each policy $\pi_i$ defines an environment $e$.


\begin{proposition} (Causal Feature Set Existence)
Under Assumption~\ref{asp: invariant} the direct causes, i.e., parents of $x^{i}_{t+1}$, define a valid support over invariant predictors, namely $S^{*}= \ $\textbf{PA($x^{i}_{t+1}$)}.
\label{prop: 1}
\end{proposition}

The proof follows directly by applying Proposition 1 of~\citet{peters2015causal} (which itself follows from construction) to each state variable indexed by dimension $i$.

The key idea therefore is to make sure that in predicting each state variable we use only its set of invariant predictors and not all state variables and actions (see Figure~\ref{fig: setup}). With this intuition, it becomes clearer why our original model learning problem is inherently tied with learning better representations, in that having access to a representation that discards excess information for each state variable (more formally, a causally invariant representation), would be more suited to learning an accurate model over and thus, at least in principle, lead to improved generalization performance across different parts of the state space. We now show that such a causally invariant representation is in fact a model-invariant abstraction. 

\begin{theorem}
\label{th: 1}
For the abstraction $\phi_{i}(x) = [x]_{S_{i}}$, where $S_{i}= \ $\textbf{PA($x^{i}_{t+1}$)}, $\phi_i$ is model-invariant. 
\end{theorem}

Proof in Appendix~\ref{app-sec: proofs}. It now becomes easy to see that sparsity in dynamics is central to what we have discussed so far, since if we do not have sparsity, the causally invariant representation trivially reduces to the original state $x$, thus resulting in no state aggregation. Next, we show that learning a transition model over a model-invariant abstraction $\phi$ and then planning over this model is optimal. 

\begin{assumption} (Concentratability Coefficient,~\citep{chen2019information})
\label{asp: concentratability}
There exists $C < \infty$ such that for any admissible distribution $\nu$, $\forall{(x, a)} \in \mathcal{X} \times \mathcal{A}, \ \ \frac{\nu(x, a)}{\mu(x, a)} < C\,. $
\end{assumption}
Here, an admissible distribution refers to any distribution that can be realized in the given CDP by following a policy for some timesteps. $\mu$ refers to the distribution from which the data is generated.

\begin{theorem} (Value bound)
\label{th: 2}
If $\phi$ is an ${\epsilon_{R}, \epsilon_{i, P}}$ approximate model-invariant abstraction on CDP $M$, and $M_\phi$ is the abstract CDP formed using $\phi$, then we can bound the loss in the optimal state action value function in both the CDPs as:
\begin{align*}
    \norm{[Q^{*}_{M_\phi}]_M - Q^{*}_{M}}_{2, \nu} &\leq \frac{\sqrt{C}}{1-\gamma} \norm{[Q^{*}_{M_\phi}]_M - \mathcal{T} [Q^{*}_{M_\phi}]_M}_{2, \mu},
\end{align*} 
\vspace{-0.35cm}
\begin{align*}
    \norm{[Q^{*}_{M_\phi}]_M-\mathcal{T} [Q^{*}_{M_\phi}]_M}_{2, \mu} &\leq \epsilon_{R} + \gamma \Big({\displaystyle \sum^{d}_{i=1} \epsilon_{i, P}}\Big) \frac{R_{\text{max}}}{(2(1 -\gamma))}.
\end{align*}
\end{theorem}
\vspace{-0.3cm}
Proof and all details surrounding the theoretical results are provided in Appendix~\ref{app-sec: proofs}. This result is important because we can follow this with standard sample complexity arguments that will have a logarithmic in $|\phi|$ dependence, thus guaranteeing that learning in this abstract MDP is faster and only incurs the above described sub-optimality. 

\section{Linear Case: Certainty Equivalence}
\label{sec: proof-of-concept}
\begin{figure*}[t]
    \hspace{-0.7cm}
    \includegraphics[width=1.1\linewidth]{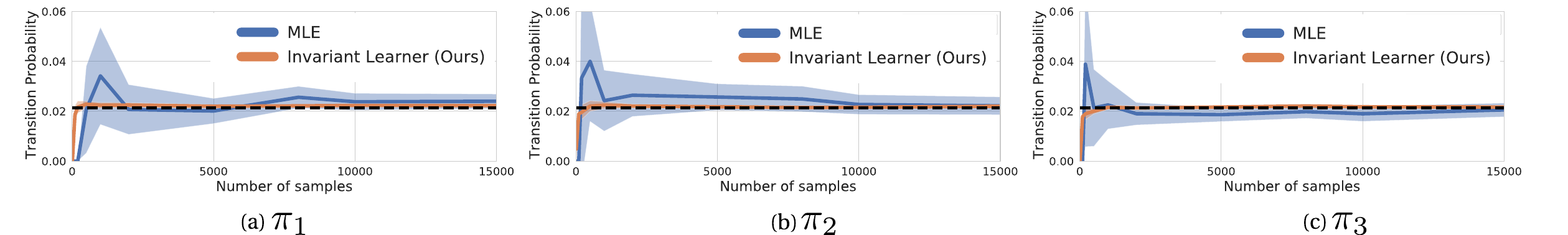}
    \caption{Consider the simple linear MDP from Appendix~\ref{subsec: linear-mdp}. We compare the mean and std. err. over 10 random seeds of the estimated transition probability of our invariant learner (\textcolor{orange}{orange curve}) and MLE (\textcolor{blue}{blue curve}). Our invariant learner converges faster and more stably to the solution (dashed black curve).}
    \label{fig: erm_irm}
    \vspace{0.2cm}
\end{figure*}

In the tabular case, estimating the model using transition samples and then planning over the learned model is referred to as certainty equivalence \citep{bertsekas1995dp}. Particularly for estimating the transition model, it considers the case where we are provided with $n$ transition samples per state-action pair, $(x_t, a_t)$ in the dataset $D_{x, a}$, and estimate the model as
\begin{align}
\label{eq: model-ERM}
    P(x_{t+1} | x_t, a_t) = \frac{1}{n} \sum_{\bar{x} \in D_{x, a}} \I(\bar{x} = x_{t+1})\,.
\end{align}
This refers to the standard maximum likelihood learner (MLE). Now, if we assume that the next state components do not depend on each other given the previous state and action (i.e., Assumption~\ref{asp: transition factorization}), we can re-write $P(x_{t+1} | x_t, a_t)$ as $\prod_{i} P(x_{t+1}^{i} | x_t, a_t)$. Assuming we know the parents of $x^{i}_{t+1}$, we can instead empirically estimate the true transition probabilities as
\vspace{-0.2cm}
\begin{align}
\label{eq: model-IRM}
    P(x_{t+1}^{i} | x_t, a_t) &= P(x_{t+1}^{i} | \textbf{PA}(x_{t+1}^{i}), a_t)
    = \frac{1}{nk} \sum_{\bar{x} \in D} \I(\bar{x} = x_{t+1}^{i})\,,
\end{align}
where $D = {\displaystyle \bigcup_{i=1}^{k}} D_{x, a}, \ x \in \phi_{i}^{-1}(\bar{x})$.
In the tabular case, Eq.~\ref{eq: model-ERM} corresponds to a solution obtained by a standard MLE learner. On the other hand, Eq.~\ref{eq: model-IRM} corresponds to a solution obtained by an invariant model learner. Proposition~\ref{prop: 1} showed that such an invariant solution exists for the given causal abstraction definition. From Theorem~\ref{th: 1}, we know that such an invariant solution is defined by the causal parents of each state variable. Therefore, in the linear dynamics case (Definition~\ref{def: causal-setup}), given data from multiple environments (different policies), we can use ICP to learn the causal parents of each state variable and then estimate the probability of a certain transition using Eq.~\ref{eq: model-IRM}. Using Theorem~\ref{th: 2}, we know that performing planning over this estimated dynamics model, for a given reward function, would be $\epsilon$-optimal. We refer to such an algorithm as model-invariant MBRL for the linear dynamics case (see Algorithm~\ref{alg: Invariant-MBRL-Linear}). Furthermore, the invariance based solution performs zero shot generalization to unseen parts of the state space while the standard model learner does not. To see this, consider a simple linear MDP with three state variables ($x^1, x^2, x^3$), each depending only on its own ($\textbf{PA}(x_{t+1}^{i}) = x_t^i$), and taking integer values between $[-10, 10]$. The exact details of the MDP are described in Appendix~\ref{subsec: linear-mdp}. We consider three different distributions corresponding to three different policies, each describing an ICP environment. In Figure~\ref{fig: erm_irm}, we compare our invariant learner (\textcolor{orange}{orange curve}) against a standard MLE learner (\textcolor{blue}{blue curve}) for the three environments/policies and show how their estimate varies as the number of samples grows.

\begin{wrapfigure}{L}{0.5\textwidth}
\begin{minipage}{0.5\textwidth}
\vspace{-25pt}
\begin{algorithm}[H] 
\begin{small}
\caption{Linear Model-Invariant MBRL}
\label{alg: Invariant-MBRL-Linear}
\begin{algorithmic}[1]
\STATE {\bfseries Input} Replay buffer $\mathcal{D}$ containing data from multiple policies/envs, confidence parameter $\alpha$; 
\FOR{state variable $i=1, \ldots, d$}
    \STATE $\mathcal{S}_i \leftarrow \texttt{ICP}( i, \mathcal{D}, \alpha)$; 
    \STATE Estimate $\hat{P}_i$ from $\mathcal{D}$ using Eq.~\ref{eq: model-IRM};
\ENDFOR
\STATE Estimate transition probability kernel $\hat{P} \leftarrow \Pi_i \hat{P}_i$
\FOR{state $ s_j, \ j =1,\ldots,N$}
    \STATE $\pi_R(s_j) \leftarrow \texttt{Plan}(R, \hat{P}, s_j)$
\ENDFOR
\end{algorithmic}
\end{small}    
\end{algorithm}
\end{minipage}
\vspace{-10pt}
\end{wrapfigure}
Note that each environment in Figure~\ref{fig: erm_irm} is specified by a fixed policy that is used for data collection. If the policy changes, it would result in a different environment as described in \cref{subsec: inv_intro}. Our ideal scenario is to find a predictive model that is closest to the true model for all environments. We find that the invariant learner quickly converges to approximately the same solution across all training environments, with just a few data samples. As can be seen, this common solution (i.e., ~0.02) also coincides with the true probability we are trying to estimate. On the other hand, the standard MLE learner results in different solutions for each training environment in the low data regime. The solution provided at test time in such a case is an average of all such solutions found during training, which is clearly off the true probability.

It is worth noting that this example assumes linearity in dynamics, which allows us to use the causal tools from~\citet{peters2015causal}. In the general non-linear case, this is not possible. To that end, in the next section we will describe a practical method that leverages self-supervised learning ideas to exploit sparsity in an end-to-end manner.

\section{Non-linear Case: Learning Practical Model-Invariant Representations}
\label{sec: practical-model-invariance}
We now introduce a practical algorithm for learning model-invariant representations wherein we relax the following assumptions: linearity in dynamics, having access to data from multiple environments, and strictly being in the batch setting. Since we wish to come up with representations that abstract away irrelevant information on a \textit{per-state-variable} level and learn them in an end-to-end manner, we propose a method which only approximates the framework described thus far. In particular, we view the task of learning such representations as a self-supervised method where we want to be invariant to models that exhibit spurious correlations. The key idea is to view two randomly initialized and independently trained (on different samples) dynamics models as augmented versions of the \textit{true} dynamics models. Since both these models exhibit spurious correlations, we wish to be invariant to such augmentations, but on a per-state-variable level (since spurious correlations arise for individual state variable dynamics). 
Specifically, we instantiate two identical models at the start of training. At each optimization step, a model is sampled randomly and is used for minimizing the standard MLE model predictive loss. Simultaneously, an invariance loss defined over the predictions of both models is attached to the MLE loss. The role of the invariance loss is essentially to \textit{minimize the difference in similarities between the prediction of one model w.r.t. the predictions of the second model and vice versa} (Eq.~\ref{eq: deep-loss}). 
It is important to note that this similarity is computed for a single state variable (randomly selected) at each training step. We borrow the specifics of the similarity definition from~\citep{mitrovic2020representation} and detail out our exact implementation of the invariance loss in pseudocode form in Appendix~\ref{subsec: pseudocode}.

The overall loss used to learn the dynamics model is thus:
\begin{equation}
\begin{small}
\label{eq: deep-loss}
    \mathcal{L}_f = \mathbb{E}_{x \sim \mathcal{D}} \Big[ \underbrace{ \Big(f(x_t, a_t) - x_{t+1}\Big)^2}_{\text{Standard MLE Loss}} + \underbrace{ \text{KL}\big( \psi^{i}(f, h), \ \psi^{i}(h, f) \big) }_{\text{Invariance Loss}} \Big],
\end{small}
\end{equation}
where $\psi^{i}(f, h) = \inp[\Big]{ g(f^{i}(x_t, a_t)) } { g(h^{i}(x_t, a_t)) } $ is the similarity between the predictions for the models $f$ and $h$ for the state variable indexed by $i$. The function $g$ is a fully connected neural network that is often called the critic in self-supervised learning losses~\citep{chen2020simple}. Note that the matrix $\psi^i$ is not symmetric since the values at any two symmetric indices are different (they are computed for different samples). Hence, the KL loss remains well-defined.

 
Eventually, we wish to use the invariant model learner described above within a model-based policy optimization algorithm and compare its policy performance to a standard MLE based model learner. 
A general framework that uses an invariant model learner is outlined in Algorithm~\ref{alg: Invariant-MBRL}. For the purposes of this paper, we employ a simple actor-critic setup where the model is used to compute multi-step estimates of the $Q$~value used by the actor learner. A specific instantiation of this idea of model value expansion is the SAC-SVG algorithm proposed in~\citet{amos2020model} (see Appendix~\ref{appsec: sac-svg} for details). It is important to note that the proposed version of model-invariance can be used in combination with any MBRL method, and with any type of dynamics model architecture, such as ensembles or recurrent architectures.
\begin{algorithm}[t]
\begin{small}
\caption{Non-linear Model-Invariant MBRL}
\label{alg: Invariant-MBRL}
\begin{algorithmic}[1]
\STATE {\bfseries Input} Replay buffer $\mathcal{D}=\emptyset$; Value and policy network parameters $\theta_{Q}$, $\theta_{\pi}$ for any MBRL algorithm;
\FOR{environment steps $t =1,\ldots,T$}
    \STATE Take action $a_t\sim\pi(\cdot|x_t)$, observe $r_t$ and $x_{t+1}$, and add to the replay buffer $\mathcal{D}$;
    \FOR{$M_{\text{model-free}}$ updates}
        \STATE Sample a batch $\{(x_j, a_j, r_j, x_{j+1})\}_{j=1}^N$ from $\mathcal{D}$;
        \STATE Run gradient update for the model free components of the algorithm (e.g. $\theta_{\pi}$, $\theta_{Q}$ etc.)
    \ENDFOR
    \FOR{$M_{\text{model}}$ updates}
        \STATE Sample a batch $\{(x_j, a_j, r_j, x_{j+1})\}_{j=1}^N$ from $\mathcal{D}$;
        \STATE Update reward model ($\theta_r$)
        \STATE Update invariant dynamics model: \ $\theta_{f} \leftarrow \text{invariant\_update}(\theta_{f}, \nabla_{\theta_{f}}L_{f} )$ (Pseudocode~\ref{subsec: pseudocode})
    \ENDFOR
\ENDFOR
\end{algorithmic}
\end{small}
\end{algorithm}
\section{Experiments}
Our experiments address the following questions: 

\begin{itemize}[noitemsep,topsep=0pt,parsep=0pt,partopsep=0pt,leftmargin=*]
    \item Moving to more complex tasks than the tabular setup in Section~\ref{sec: proof-of-concept}, can we show the adverse effects of spurious correlations arising due to learning the model as the policy distribution, and thus the state distribution change during learning~(\cref{subsec: spurious-correlation})? 
    \item How does our invariant model learning scheme perform in comparison to a standard MLE based model learner on more challenging tasks? Does the performance gain, if any, have any correlation with the number of samples, i.e., amount of data available~(\cref{subsec: invariant-model-learning})?
    \item How does learning an invariant model affect downstream performance? Does learning a more accurate model result in more sample efficient algorithms~(\cref{subsec: invariant-mbrl})?
\end{itemize}

\clearpage
\subsection{Presence of Spurious Correlations}
\label{subsec: spurious-correlation}

\begin{wrapfigure}{r}{0.4\linewidth}
     \centering
     \vspace{-15pt}
     \begin{subfigure}[t]{0.18\textwidth}
         \includegraphics[scale=0.2]{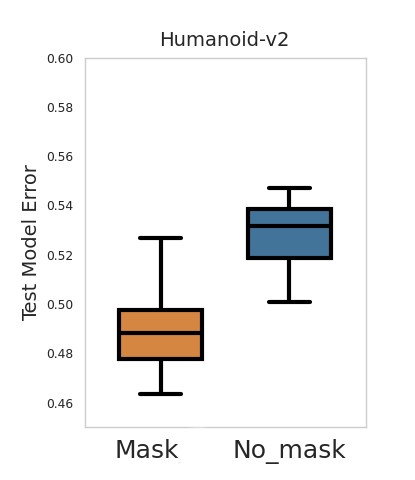}
         \caption{Horizon=3}
     \end{subfigure} 
     \begin{subfigure}[t]{0.18\textwidth}
         \includegraphics[scale=0.2]{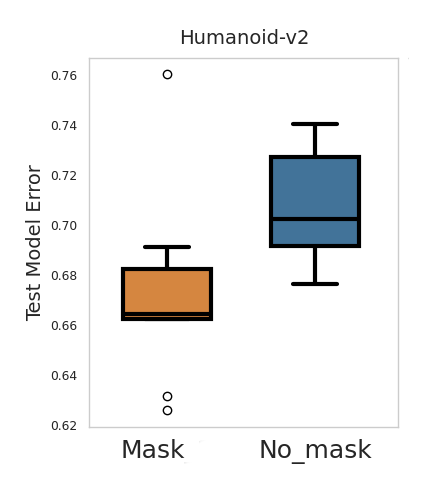}
         \caption{Horizon=5}
     \end{subfigure} 
     \caption{Effect of spurious correlation on the model learning test loss of Humanoid-v2 for a single dimension (the knee joint) with two schemes: Mask\_2 and No\_mask.  10 seeds, 1 std. dev. shaded. Y-axis magnitude order is 1e-3.}
     \vspace{-15pt}
     \label{fig: spurious-correlation}
\end{wrapfigure}


To test the presence of spurious correlations when learning the dynamics model, we present the following experiment. For the Humanoid-v2 domain, we choose to predict a single dimension (the knee joint) when {\bf 1) No\_Mask:} the entire current observation and action are provided as input and, {\bf 2) Mask:} when the dimensions that are likely uncorrelated to the knee joint are masked (see Appendix~\ref{appsec: spurious-correlation}). Having trained different models for the two cases, we observe that {\bf 1) No\_Mask:} performs worse than {\bf 2) Mask:}, for both horizon values in $\{3, 5\}$ (see Figure~\ref{fig: spurious-correlation}). This indicates that there indeed is an invariant, causal set of parents among the observation dimensions and that there could be some interference due to spurious correlations in 1), and thus, it performs worse than case 2).
Furthermore, when the dimensions that are likely to be useful in predicting the knee joint are masked, then the model error is the highest (worst; not shown in figure). 

\begin{wrapfigure}{r}{0.4\linewidth}
    \vspace{-35pt}
    \includegraphics[scale=0.25]{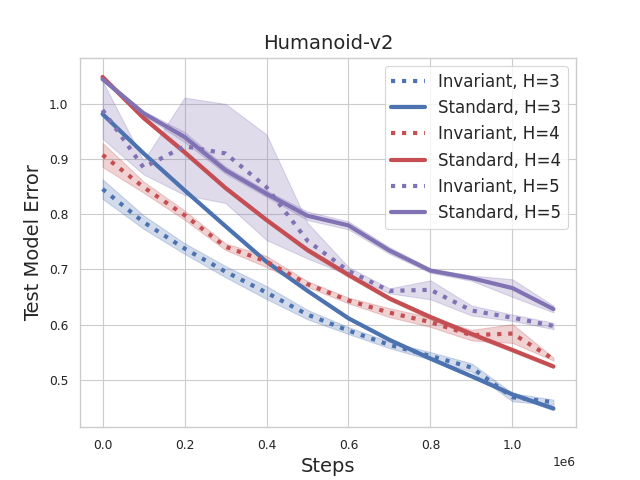}
    \caption{Test model learning error on Humanoid-v2 for different horizon values. Mean and std. err. over 10 random seeds.}
    \vspace{-12pt}
    \label{fig: model-error}
\end{wrapfigure}

\subsection{Invariant Model Learning on Humanoid-v2}
\label{subsec: invariant-model-learning}

We compare the invariant model learner to a standard model learner for the Humanoid-v2 task. To observe the effect of the invariance loss clearly, we decouple the model learning component from the policy optimization component by testing the model on data coming from a replay buffer, a pre-trained model-free SAC agent. Such a setup ensures that the change in state distribution according to changes in policy is still present, which is necessary to test the generalization performance of a learned model. 

We observe that our invariant model learner performs much better than the standard model learner, especially when the number of samples available is low, i.e., around the 200k to 500k mark (see Figure~\ref{fig: model-error}). As the number of samples increases, the performance between both models converges, just as observed in the tabular case. This is expected since in the infinite data regime, both solutions (MLE and invariance based) approach the true model. Furthermore, we observe that the number of samples it takes for convergence between the standard and the invariant model learners increases as the rollout horizon (H in Figure~\ref{fig: model-error}) of the model learner is increased.

\begin{figure}[t]
\begin{minipage}[t]{\textwidth}
\hspace{-0.07in}
\begin{minipage}{0.5\linewidth}
\resizebox{1\linewidth}{!}{%
\begin{tabular}{ 
  >{\raggedright\arraybackslash}lcccc} \toprule
POPLIN & Cheetah & Walker  & Hopper & Ant  \\ \hline \vspace{-0.2cm} \\
 
 *PETS &  2288 {\tiny$\pm$510}  & 282 {\tiny$\pm$ 250} & 114 {\tiny$\pm$ 311} & 1165 {\tiny$\pm$ 113}   \\
 
 *POPLIN-A &  1562 {\tiny$\pm$568}  & -105 {\tiny$\pm$ 125} & 202 {\tiny$\pm$ 481} & 1148 {\tiny$\pm$ 219}   \\
 
 *POPLIN-P & 4235 {\tiny$\pm$566}  & {\bf 597} {\tiny$\pm$ 239} & {\bf 2055} {\tiny$\pm$ 206} & 2330 {\tiny$\pm$ 160}   \\
 
 $^{*}$METRPO &  2283 {\tiny$\pm$ 450}  & -1609 {\tiny$\pm$328} & 1272 {\tiny$\pm$ 250} & 282 {\tiny$\pm$ 9}   \\
 
 *SAC &  4035 {\tiny$\pm$ 134}  & -382 {\tiny$\pm$ 424} & 2020 {\tiny$\pm$ 346} & 836 {\tiny$\pm$ 34}  \\
 
 SAC-SVG H-3 &  8211 {\tiny$\pm$408}  & -242 {\tiny$\pm$ 606} & 1869 {\tiny$\pm$389} & 3977 {\tiny$\pm$ 357}  \\
 
 Ours H-3 &  {\bf 8509} {\tiny$\pm$470} & -768 {\tiny$\pm$ 427} & 1801 {\tiny$\pm$ 355} & {\bf 4521} {\tiny$\pm$ 307}  \\
 
 \bottomrule
\end{tabular}
}
\captionof{table}{}
\label{tab: invariant-mbrl-poplin}
\end{minipage}
\hspace{0.2cm}
\begin{minipage}{0.5\linewidth}
\resizebox{\linewidth}{!}{%
\begin{tabular} { 
  >{\raggedright\arraybackslash}lcccc} \toprule
MBPO & Cheetah & Walker  & Hopper & Ant \\ 
 \hline \vspace{-0.2cm} \\
 SAC-SVG H-3 & {\bf 7296} {\tiny$\pm$462}  & {\bf 3274} {\tiny$\pm$364} & 3055 {\tiny$\pm$ 91} & 3090 {\tiny$\pm$ 160} \\ 
 
 \vspace{0.05cm}
 Ours H-3 & 7253 {\tiny$\pm$395} & 2882 {\tiny$\pm$416} & {\bf 3090} {\tiny$\pm$ 109} & {\bf 3424} {\tiny$\pm$ 320} \\ 
 
 \hline \vspace{-0.2cm} \\
 SAC-SVG H-4 & 6917 {\tiny$\pm$564}  & 3190 {\tiny$\pm$374} & 3109 {\tiny$\pm$1126} & 828 {\tiny$\pm$ 435}  \\ 
 
 \vspace{0.05cm}
 Ours H-4 & {\bf 7206} {\tiny$\pm$ 327}  & {\bf 3392} {\tiny$\pm$407} & {\bf 3204} {\tiny$\pm$ 115} & {\bf 2222} {\tiny$\pm$383} \\ 
 
 \hline \vspace{-0.2cm} \\
 SAC-SVG H-5 & 4305 {\tiny$\pm$1025}  & 2538 {\tiny$\pm$492} & 2820 {\tiny$\pm$ 316} & {\bf 2440} {\tiny$\pm$479} \\ 
 
 Ours H-5 & {\bf 6602} {\tiny$\pm$ 345}  & {\bf 2916} {\tiny$\pm$442} & {\bf 3009} {\tiny$\pm$ 280} & 2162 {\tiny$\pm$490} \\ 
 \bottomrule
\end{tabular}
}
\captionof{table}{}
\label{tab: invariant-mbrl-mbpo}
\end{minipage}
\end{minipage}
\caption*{Table : Invariant MBRL performance reported at 200k steps on four MuJoCo based domains from POPLIN~\citep{wang2019exploring} and on five MuJoCo based domains from MBPO~\citep{janner2019trust}. * represents performance reported by POPLIN. Standard error with 10 seeds reported. We bold the scores with larger mean values.}
\vspace{0.2cm}
\end{figure}

\subsection{Invariant Model-based Reinforcement Learning}
\label{subsec: invariant-mbrl}
Finally, we evaluate the invariant model learner within the the policy optimization setting of SAC-SVG~\citep{amos2020model}. We compare the difference in performance to SAC-SVG when the horizon length is varied (see MBPO environments in Table~\ref{tab: invariant-mbrl-mbpo} and Appendix~\ref{appsec: mbpo-envs}) and then compare the performance of our method against multiple model based methods including PETS~\citep{chua2018deep}, POPLIN~\citep{wang2019exploring}, METRPO~\citep{kurutach2018model}, and the model free SAC~\citep{haarnoja2018soft} algorithm (see POPLIN environments in Table~\ref{tab: invariant-mbrl-poplin} and Appendix~\ref{appsec: mbpo-envs}). The results show improved performance when the invariant model learner is used instead of the standard model learner across most tasks. Furthermore, we see that as the horizon lengt h is increased, the difference in performance between the invariant and standard learners increases for most tasks. Combining our invariant model learner with other policy optimization algorithms is therefore a promising direction for future investigation.

\section{Related Work}
\textbf{On Factored MDPs}: Planning based on structural assumptions on the underlying MDP have been explored in significant detail in the past~\citep{boutilier1999decision}. The most closely related setting is of factored MDPs, but learning based approaches that build on the factored MDP assumption have predominantly also assumed a known graph structure for the transition factorization \citep{kearns1999efficient, strehl2007efficient, osband2014near}.
On the theory side, most prior works on factored MDPs also do not learn and leverage state abstractions \citep{kearns1999efficient, strehl2007efficient}. \citet{jonsson2006causalgraph} draw connections to causal inference, but do so explicitly with dynamic Bayesian networks (DBN)--- as opposed to learning approximate abstractions--- and assume knowledge of the model.
Most recently, \citet{misra2021provable} tackle the rich observation factored MDP setting, but consider each pixel an atom that belongs to a single factor. 
On the algorithmic side, there have been only a few works that discuss learning the graph or DBN structure alongside the factored MDP assumption, e.g.,~\citet{hallak2015off}. We differ from these in that we only learn the partial graph structure (not explicitly), i.e., only the direct parents of each state variable. Moreover, we achieve this using the invariance principle, which has not been explored in prior work. Finally, most factored MDP works include the factored reward condition as well and do not propose a practical method that works for complex environments like the ones based on MuJoCo, often sticking more to theoretical contributions.


\textbf{On CDPs}: There has been a lot of recent work around the newly proposed CDP setting. Our work has overlapping ideas with two specific works --- model based learning in CDPs \citep{misra2020kinematic} and learning efficient abstractions over them \citep{sun2019model}. Besides the more algorithmic and empirically focused nature of this work, there remain several considerable distinctions. Firstly, we focus on abstraction-based learning, whereas \citet{sun2019model} rely on the concept of \textit{witness misfit} to learn efficiently over the original CDP states. Secondly, we are focused on learning abstract states that are a coarser representation of the true full state of the CDP, whereas \citet{misra2020kinematic} deal with the case where the abstract states correspond to the mapping from rich observation to the full state of the CDP. In that sense, the framework adopted here is a blend of that presented in these two works. Ideally, we would like to show that the class of problems where the number of model-invariant abstract states is low also has a low \textit{witness rank}. 

\section{Conclusion and Future Directions}

This paper introduced a new type of state abstraction for MBRL that exploits the inherent sparsity present in many complex tasks. We first showed that a representation that only depends on the causal parents of each state variable follows this definition and is provably optimal. Following, we introduced a novel approach for learning model-invariant abstractions in practice, which can be plugged in any given MBRL method. Experimental results show that this approach measurably improves the generalization ability of the learnt models. This stands as an important first step to building more advanced algorithms with improved generalization for systems that possess sparse dynamics.

In terms of future work, there remain multiple exciting directions and open questions. First, to enable model-invariance, we could also look at other kind of approaches proposed recently such as the AND mask \citep{parascandolo2020learning}. The AND mask specifically requires the data to be separated into multiple environments, and thus naturally suits the offline RL setting. Second, moving to pixel based input, the representation learning task becomes two-fold, including learning to abstract away the irrelevant information present in the pixels and then learning a model-invariant representation. Third, note that our theoretical results do not involve an explicit dependence on a sparsity measure, for example, the maximum number of parents any state variable could have. Including such a dependence would ensure tighter bounds. Fourth, it is worth asking how such an explicit constraint on model-invariance can perform as a standalone representation learning objective, considering the strong progress made by self-supervised RL.

\bibliography{NeurIPS}
\bibliographystyle{icml2021}

\clearpage
\section*{Checklist}


\begin{enumerate}

\item For all authors...
\begin{enumerate}
  \item Do the main claims made in the abstract and introduction accurately reflect the paper's contributions and scope?
    \answerYes{}
  \item Did you describe the limitations of your work?
    \answerYes{}
  \item Did you discuss any potential negative societal impacts of your work?
    \answerNA{}
  \item Have you read the ethics review guidelines and ensured that your paper conforms to them?
    \answerYes{}
\end{enumerate}

\item If you are including theoretical results...
\begin{enumerate}
  \item Did you state the full set of assumptions of all theoretical results?
    \answerYes{}
	\item Did you include complete proofs of all theoretical results?
    \answerYes{}
\end{enumerate}

\item If you ran experiments...
\begin{enumerate}
  \item Did you include the code, data, and instructions needed to reproduce the main experimental results (either in the supplemental material or as a URL)?
    \answerYes{}
  \item Did you specify all the training details (e.g., data splits, hyperparameters, how they were chosen)?
    \answerYes{}
	\item Did you report error bars (e.g., with respect to the random seed after running experiments multiple times)?
    \answerYes{}
	\item Did you include the total amount of compute and the type of resources used (e.g., type of GPUs, internal cluster, or cloud provider)?
    \answerNo{}
\end{enumerate}

\item If you are using existing assets (e.g., code, data, models) or curating/releasing new assets...
\begin{enumerate}
  \item If your work uses existing assets, did you cite the creators?
    \answerYes{}
  \item Did you mention the license of the assets?
    \answerNo{}
  \item Did you include any new assets either in the supplemental material or as a URL?
    \answerNo{}
  \item Did you discuss whether and how consent was obtained from people whose data you're using/curating?
    \answerNA{}
  \item Did you discuss whether the data you are using/curating contains personally identifiable information or offensive content?
    \answerNA{}
\end{enumerate}

\item If you used crowdsourcing or conducted research with human subjects...
\begin{enumerate}
  \item Did you include the full text of instructions given to participants and screenshots, if applicable?
    \answerNA{}
  \item Did you describe any potential participant risks, with links to Institutional Review Board (IRB) approvals, if applicable?
    \answerNA{}
  \item Did you include the estimated hourly wage paid to participants and the total amount spent on participant compensation?
    \answerNA{}
\end{enumerate}

\end{enumerate}


\def\thesection{\Alph{section}}
\clearpage
\newpage

\xpretocmd{\part}{\setcounter{section}{0}}{}{}
\xpretocmd{\part}{\setcounter{table}{0}}{}{}
\xpretocmd{\part}{\setcounter{theorem}{0}}{}{}
\xpretocmd{\part}{\setcounter{lemma}{0}}{}{}

\part*{Appendix}


\section{Why Causal Invariance?}
\label{sec: Motivation}

Out of distribution (OOD) generalization has been attributed to learnt correlations that do not follow the underlying causal structure of the system. These are referred to as spurious correlations. With the use of deep neural networks, spurious correlations can arise due to 1) the way we collect data, or selection bias, 2) overparameterization of the neural networks, and 3) presence of irrelevant information in the data (ex. the background might be irrelevant for an object classification task). For the setting in this paper, such issues are relevant since we use NNs to learn the dynamics model of the RL environment. Even if these issues are attended to, spurious correlation could still arise. However, this time it would be due to the causal structure assumed and not the modelling technique (NNs) we use over it. Two such causes are 4) hidden confounders in the causal graph and 5) conditioning on anti-causal parts of input $x$. For our case, 4) could correspond to a hidden non-stationarity in the system such as the friction coefficient between the robot and the floor. Since we are only concerned with the $x_t$ to $x_{t+1}$ causal diagram, 5) may not be as apparent. Nevertheless, we include it for completeness. Therefore, in principle, choosing the right variables and deploying techniques that discover an invariant Y conditioned on a given X helps us avoid spurious correlations. This in turn leads to better OOD generalization.

\section*{Notes on Assumptions}

\begin{itemize}
    \item There is a linearity assumption on the dynamics that is implicitly placed when we borrow the generalization results of ~\citet{peters2015causal}. These ensure that given data divided into multiple environments (minimum 2) (in our case that refers to data from multiple single policies), the causal representation results in a model that generalizes over all environments. When the dynamics are non-linear, ~\citet{arjovsky2019invariant} showed that a similar argument toward generalization can still be made, with the added requirement of having data from at least a fixed amount ($n_e \ge 2$) of environments. However, recent work~\citep{rosenfeld2020risks} has argued that such an analysis is not accurate and thus more investigation is required to ensure OOD generalization. For the proof of concept experiment in Section~\ref{sec: proof-of-concept}, the dynamics are linear and thus we can deploy ICP for learning the causal parents of each state variable and ensure that the zero-shot generalization shown actually persists for any arbitrarily different policy from the ones used for training the invariant learner. When we move to Section~\ref{sec: practical-model-invariance} we do away with this assumption since the dynamics are no longer linear. Moreover, we do not restrict ourselves to a multiple environment based regime, the likes of which are required by ~\citet{peters2015causal}.
    \item The transition factorization assumption, i.e. Assumption~\ref{asp: transition factorization}, seems like a strict condition in theory when we move to complex domains, however, it is in fact a natural outcome of how we model the agent dynamics in practice. In practice, each state variable of the next state $x_{t+1}$ is set to only be dependent on the previous state $x_t$ and action $a_t$. We can see this for example in neural network based dynamics models where the next state as a whole (all state variables simultaneously) is predicted given the previous state and action. Therefore, even though it may seem as an over constraining assumption, in practice it is present by default. In fact, this shows that we should focus more on theoretical results that build on assumptions like transition factorization.  
    \item A constraint on the exploration issue is usually dealt with by the concentrability assumption (Assumption ~\ref{asp: concentratability}) in literature. A recent method to get around such an assumption is by coupling the policy optimization algorithm with a exploration algorithm that maintains a set of exploratory policies (\textit{policy cover} in \citet{misra2020kinematic}) which slowly keeps expanding.  
    \item When describing the practical invariant model learner (Section~\ref{sec: practical-model-invariance}), we do not explicitly focus on finding the exact causal parents for each state variable. On the other hand, we resort to forcing such a constraint implicitly by describing a direct, differentiable invariance-based loss. One benefit of this approach is that the overall method remains end-to-end. The downside of course is that we do not always ensure that the right set of causal parents is found.  
\end{itemize}

\newpage
\section{Proofs}
\label{app-sec: proofs}

\begin{theorem}
\label{eq: theorem1}
For the abstraction $\phi_{i}(x) = [x]_{S_{i}}$, where $S_{i}= \ $\textbf{PA($x^{i}_{t+1}$)}, $\phi_i$ is model-invariant. Furthermore, if $\phi$ follows such a definition for all state variables indexed by $i$, $\phi$ is a reward free model irrelevant state abstraction. 
\end{theorem}

\emph{Proof.} We first prove that $\phi_i$ is model-invariant. In the case where $\phi_{i}(x) = \phi_{i}(x^\prime)$ for some state variable indexed by $i$, we have:

\begin{align*}
    P(\bar{x}^i | x, a) &=  P(\bar{x}^i | [x]_{S_{i}}, a)  \\
    &= P(\bar{x}^i | \phi_{i}(x), a)  \\
    &= P(\bar{x}^i | \phi_{i}(x^\prime), a).
\end{align*}

Following the same steps backwards for $\phi_i(x^\prime)$ concludes the proof.

We now prove the latter statement in the theorem. Note that for such a statement to be meaningful, we require that the state space $\mathcal{X}$ includes some irrelevant state variables for the downstream task in hand. For example, we could have some unnecessary noise variables appended to the full state variables. In such a case, the full state variables are relevant for the downstream task whereas the noise variables are irrelevant for the downstream task. Now, if $\phi(x) = \phi(x^\prime)$, i.e., $\phi_i(x) = \phi_i(x^\prime)$ for all relevant state variables indexed by $i$, $\phi$ is a reward free model irrelevant state abstraction, i.e.,

\begin{align}
    \sum_{\bar{x} \in \phi^{-1}(s)} P(\bar{x} | x, a) = \sum_{\bar{x} \in \phi^{-1}(s)} P(\bar{x} | x^{\prime}, a),
\end{align}

where $s$ is the abstract state that $\phi$ maps to. With this note, the proof for the latter statement follows directly from Theorem 1 in \citet{zhang2020invariant}. 

\textbf{On the absence of irrelevant state variables:} The condition $\phi(x^1) = \phi(x^2)$ is quite strict if we assume the absence of irrelevant state variables (if no such variables are present, then $x^1$ has to be equal to $x^2$ for this condition to be met, which is not meaningful). 

\textbf{Extending to model-invariance grounded in reward:} Notice that Definition~\ref{def: model-invariant-abstraction} is reward free, and is grounded in the next state $\bar{x}$. We could instead extend this to a definition which is grounded in the reward. Particularly,

\begin{definition} (Reward Grounded Model Invariant Abstraction)
\label{def: reward-model-invariant-abstraction}
$\phi_i$ is reward grounded model-invariant if for any $x, x^{\prime},\bar{x} \in \mathcal{X},  a \in \mathcal{A}$, $\phi_{i}(x) = \phi_{i}(x^{\prime})$ if and only if 
\begin{align*}
        R_i(x, a) &= R_i(x^\prime, a) \\
       \sum_{\bar{x} \in \phi_i^{-1}(s_i)} P(\bar{x} | x, a) &= \sum_{\bar{x} \in \phi_i^{-1}(s_i)}  P(\bar{x} | x^{\prime}, a),
\end{align*}
\end{definition}

We can show that the causal representation of $\phi$ is a reward free version of the above defined model-invariance abstraction (Definition~\ref{def: reward-model-invariant-abstraction}).

\begin{proposition}
For the abstraction $\phi_{i}(x) = [x]_{S_{i}}$, where $S_{i}= \ $\textbf{PA($x^{i}_{t+1}$)}, $\phi_i$ is a reward free version of Definition~\ref{def: reward-model-invariant-abstraction}.
\end{proposition}

\emph{Proof.}
Now, when $\phi_{i}(x) = \phi_{i}(x^\prime)$ for a specific state variable indexed by $i$, we have:

\begin{align*}
    {\displaystyle \sum_{\bar{x} \in \phi_{i}^{-1}(s)}} P(\bar{x} | x, a) &= {\displaystyle \sum_{\bar{x} \in \phi_{i}^{-1}(s)}} \         {\displaystyle \prod_{k=1}^{d}} P(\bar{x}^{k} | x, a) \\
    &= {\displaystyle \sum_{\bar{x} \in \phi_{i}^{-1}(s)}} P(\bar{x}^i | [x]_{S_{i}}, a) \  {\displaystyle \prod_{k=1}^{d}} P(\bar{x}^{k \neq i, \ i, k \in [d]} | x, a) \\
    &= f(s, \phi_{i}(x), a) {\displaystyle \sum_{\bar{x} \in \phi_{i}^{-1}(s)}} \ P(\{\bar{x}\}^{k \neq i, \ i, k \in [d]} | x, a) \\
    &= f(s, \phi_{i}(x), a)  \\
    &= f(s, \phi_{i}(x^\prime), a).
\end{align*}

, where we use $[d]$ to denote the set $\{ 1, \cdots, d\}$ and $f$ is some function that depends only on the abstraction $\phi_i$, action $a$, and the abstract state $s$. Following the same steps backwards concludes the proof.


\begin{lemma} (Model Error Bound)
Let $\phi$ be an $\epsilon_{i,P}$-approximate model-invariant abstraction on CDP $M$. Given any distributions ${p_{s_i} : s_i \in \phi_i(\mathcal{X})}$ where $p_{s_i}$ is supported on $\phi^{-1}({s_i})$ and $p_s = \prod^{d}_{i=1} p_{s_i}$, we define $M_{\phi} = ({\phi(\mathcal{X})}, \mathcal{A}, P_{\phi}, R_{\phi}, \gamma)$ where $P_{\phi}(s, a) = \mathbb{E}_{x \sim p_s} [P(\cdot | x, a)] $. Then for any $x \in \mathcal{X}$, $a \in \mathcal{A}$,
\begin{align*}
    \norm{P_{\phi}(s, a) - \Phi P(x, a)} &\leq {\displaystyle \sum^{d}_{i=1}} \epsilon_{i, P}.
\end{align*}
\end{lemma}

\emph{Proof.} Consider any $x$, $a$ and let $q_{x^i} := \Phi_{i} P(x, a)$, where we have $\norm{q_{x^i} - q_{\bar{x}^i}} \leq \epsilon_{i, P}$ if $\phi_{i}(x) = \phi_{i}(\bar{x})$.

\begin{align*}
    \norm{P_{\phi}(s, a) - \Phi P(x, a)} &= \norm{\displaystyle \sum_{\bar{x} \in \phi^{-1}(s)} p_{s}(\bar{x}) P(\cdot | \bar{x}, a) - \Phi P(x, a)} \\
    &= \norm{\displaystyle \sum_{\bar{x} \in \phi^{-1}(s)} p_{s}(\bar{x}) P(\cdot | \bar{x}, a) - \displaystyle \prod_{i=1}^{d} \Phi_i P(x, a)} \\
    &= \norm{\displaystyle \sum_{\bar{x} \in \phi^{-1}(s)} p_{s}(\bar{x}) \prod_{i=1}^{d} q_{\bar{x}^i} -  \prod_{i=1}^{d} q_{x^i}} \\
    &= \norm{\displaystyle \sum_{\bar{x} \in \phi^{-1}(s)} p_{s}(\bar{x}) \Big( \prod_{i=1}^{d} q_{\bar{x}^i} - \prod_{i=1}^{d} q_{x^i}\Big)} \\
    &\leq \displaystyle \sum_{\bar{x} \in \phi^{-1}(s)} p_{s}(\bar{x}) \norm{\prod_{i=1}^{d} q_{\bar{x}^i} - \prod_{i=1}^{d} q_{x^i}}.
\end{align*}

We now use the following inequality:

\begin{align*}
    \norm{AB - CD} &= \norm{AB - AD + AD - CD} \\
    &= \norm{A(B - D) + (A - C)D} \\
    &\leq \norm{A(B - D) + \norm{(A - C)D}}  &\quad \text{(Triangle inequality)} \\
    &\leq \norm{A}_{\infty}\norm{B - D}_{1} + \norm{A - C}_{1}\norm{D}_{\infty} &\quad \text{(Holder's inequality)}. 
\end{align*}

The $\infty-$norm of a probability distribution is $1$. Apply this result to the above expression $d$ times,
\begingroup\setlength{\jot}{-0.5ex}
\begin{align*}
    \norm{P_{\phi}(x, a) - \Phi P(x, a)} &\leq \displaystyle \sum_{\bar{x} \in \phi^{-1}(s)} p_{s}(\bar{x}) \norm{\prod_{i=1}^{d} q_{\bar{x}^i} - \prod_{i=1}^{d} q_{x^i}} \\
    &\leq \displaystyle \sum_{\bar{x} \in \phi^{-1}(s)} p_{s}(\bar{x}) \Bigg( \norm{\prod_{i=1}^{d-1} q_{\bar{x}^i}}_{\infty} \norm{q_{\bar{x}^{d}} - q_{x^{d}}}_{1} + \norm{\prod_{i=1}^{d-1} q_{\bar{x}^i} - \prod_{i=1}^{d-1} q_{x^i} }_{1} \norm{q_{x^d}}_{\infty} \Bigg) \\
    & \vdots \\
    &\leq \displaystyle \sum_{\bar{x} \in \phi^{-1}(s)} p_{s}(\bar{x}) {\displaystyle \sum^{d}_{i=1}} \epsilon_{i, P} \\
    &= {\displaystyle \sum^{d}_{i=1}} \epsilon_{i, P}.
\end{align*}
\endgroup

\begin{theorem} (Value bound)
If $\phi$ is an ${\epsilon_{R}, \epsilon_{i, P}}$ approximate model-invariant abstraction on CDP $M$, and $M_\phi$ is the abstract CDP formed using $\phi$, then we can bound the loss in the optimal state action value function in both the CDPs as:
\begingroup\setlength{\jot}{0ex}
\begin{align*}
    \norm{[Q^{*}_{M_\phi}]_M - Q^{*}_{M}}_{2, \nu} &\leq \frac{\sqrt{C}}{1-\gamma} \norm{[Q^{*}_{M_\phi}]_M - \mathcal{T} [Q^{*}_{M_\phi}]_M}_{2, \mu} \\
    \norm{[Q^{*}_{M_\phi}]_M - \mathcal{T} [Q^{*}_{M_\phi}]_M}_{2, \mu} &\leq \epsilon_{R} + \gamma \Big({\displaystyle \sum^{d}_{i=1} \epsilon_{i, P}}\Big) R_{\text{max}} / (2(1 -\gamma))
\end{align*}
\endgroup
\end{theorem}
Note that this theorem deals with the batch setting, where we are given a batch of data and are tasked at learning only using this data, without allowing any direct interaction with the CDP. We use the concentratability coefficient as defined in Assumption~\ref{asp: concentratability}, i.e., there exists a $C$ such that for any admissible distribution $\nu$:
\begin{align*}
    \forall{(x, a)} \in \mathcal{X} \times \mathcal{A}, \ \ \frac{\nu(x, a)}{\mu(x, a)} < C\,. 
\end{align*}

Here, we abuse $\mu$ to represent the distribution the data comes from instead of standard notation representing the starting state distribution. Now,

\begin{align*}
    \norm{[Q^{*}_{M_\phi}]_M - Q^{*}_{M}}_{2, \nu} &= \norm{[Q^{*}_{M_\phi}]_M - \mathcal{T} [Q^{*}_{M_\phi}]_M + \mathcal{T} [Q^{*}_{M_\phi}]_M - Q^{*}_{M}}_{2, \nu} \\
    &\le \norm{[Q^{*}_{M_\phi}]_M - \mathcal{T} [Q^{*}_{M_\phi}]_M}_{2, \nu} + \norm{\mathcal{T} [Q^{*}_{M_\phi}]_M - \mathcal{T} Q^{*}_{M}}_{2, \nu} \\
    &\le \sqrt{C} \norm{[Q^{*}_{M_\phi}]_M - \mathcal{T} [Q^{*}_{M_\phi}]_M}_{2, \mu} + \norm{\mathcal{T} [Q^{*}_{M_\phi}]_M - \mathcal{T} Q^{*}_{M}}_{2, \nu} &\quad (\text{3})\\
\end{align*}

Let us consider the second term:
\begingroup\setlength{\jot}{0ex}
\begin{align*}
    \norm{\mathcal{T} [Q^{*}_{M_\phi}]_M - \mathcal{T} Q^{*}_{M}}_{2, \nu}^2 &= \mathbb{E}_{(x, a)\sim\nu} \Bigg[ \Big( \mathcal{T} [Q^{*}_{M_\phi}]_M (x, a) - \mathcal{T} Q^{*}_{M} (s, a) \Big)^2 \Bigg] \\
    &= \mathbb{E}_{(x,a)\sim\nu} \Bigg[ \Big( \gamma\mathbb{E}_{x'\sim P(x, a)} [ \max_{a} [Q^{*}_{M_\phi}]_M (x', a) - \max_{a} Q^{*}_{M} (x', a) ] \Big)^2 \Bigg] \\
    &\le \mathbb{E}_{(x,a)\sim\nu} \Bigg[  \gamma^2\mathbb{E}_{x'\sim P(x, a)} \Big( \max_{a} [Q^{*}_{M_\phi}]_M (x', a) - \max_{a} Q^{*}_{M} (x', a) \Big)^2 \Bigg] \\
    &\le \gamma^2 \mathbb{E}_{(x,a)\sim\nu} \ \mathbb{E}_{x'\sim P(x, a)} \Big[ \max_{a} \Big( [Q^{*}_{M_\phi}]_M (x', a) - Q^{*}_{M} (x', a) \Big)^2 \Big] \\
    &\le \max_{\nu} \Bigg[ \gamma^2 \mathbb{E}_{(x,a)\sim\nu} \ \mathbb{E}_{x'\sim P(x, a)} \Big[ \max_{a} \Big( [Q^{*}_{M_\phi}]_M (x', a) - Q^{*}_{M} (x', a) \Big)^2 \Big] \Bigg] \\
    &\le \max_{\nu} \Bigg[ \gamma^2 \mathbb{E}_{(x,a)\sim\nu} \Big[ \Big( [Q^{*}_{M_\phi}]_M (x', a) - Q^{*}_{M} (x', a) \Big)^2 \Big] \Bigg] \\
    &= \max_{\nu} \gamma^2 \norm{ [Q^{*}_{M_\phi}]_M - Q^{*}_{M}}_{2, \nu}^2
\end{align*}
\endgroup

where the last inequality follows because the two terms inside the expectation only depend on the next state $x'$ and the next action $a$ which can only be less than the value for $x, a \sim \nu$ since we maximize over it. 

Plugging this back in (3):
\begingroup\setlength{\jot}{0ex}
\begin{align*}
    \norm{[Q^{*}_{M_\phi}]_M - Q^{*}_{M}}_{2, \nu} 
    &\le \sqrt{C} \norm{[Q^{*}_{M_\phi}]_M - \mathcal{T} [Q^{*}_{M_\phi}]_M}_{2, \mu} + \max_{\nu} \gamma \norm{ [Q^{*}_{M_\phi}]_M - Q^{*}_{M}}_{2, \nu} \\
    \max_{\nu} \norm{ [Q^{*}_{M_\phi}]_M - Q^{*}_{M}}_{2, \nu} &\le \sqrt{C} \norm{[Q^{*}_{M_\phi}]_M - \mathcal{T} [Q^{*}_{M_\phi}]_M}_{2, \mu} + \max_{\nu} \gamma \norm{ [Q^{*}_{M_\phi}]_M - Q^{*}_{M}}_{2, \nu} \\
    \max_{\nu} \norm{ [Q^{*}_{M_\phi}]_M - Q^{*}_{M}}_{2, \nu} &\le \frac{\sqrt{C}}{1-\gamma} \norm{[Q^{*}_{M_\phi}]_M - \mathcal{T} [Q^{*}_{M_\phi}]_M}_{2, \mu} \\
\end{align*}
\endgroup
Since $\norm{ [Q^{*}_{M_\phi}]_M - Q^{*}_{M}}_{2, \nu} \le \max_{\nu} \norm{ [Q^{*}_{M_\phi}]_M - Q^{*}_{M}}_{2, \nu}$, we have:
\begin{align*}
    \norm{ [Q^{*}_{M_\phi}]_M - Q^{*}_{M}}_{2, \nu} &\le \frac{\sqrt{C}}{1-\gamma} \norm{[Q^{*}_{M_\phi}]_M - \mathcal{T} [Q^{*}_{M_\phi}]_M}_{2, \mu}
\end{align*}

Now, we prove the second statement:
\begingroup\setlength{\jot}{0ex}
\begin{align*}
    \norm{[Q^{*}_{M_\phi}]_M - \mathcal{T} [Q^{*}_{M_\phi}]_M}_{2, \mu} &\le \norm{[Q^{*}_{M_\phi}]_M - \mathcal{T} [Q^{*}_{M_\phi}]_M}_{\infty} \\
    &= \norm{[\mathcal{T}_{M_\phi} Q^{*}_{M_\phi}]_M - \mathcal{T} [Q^{*}_{M_\phi}]_M}_{\infty} \\
    &= \sup_{x,a} |R_\phi(\phi(x), a) + \gamma \ip{P_\phi(\phi(x), a), V^*_{M_\phi}} - R(x, a) - \gamma \ip{P(x,a), [V^*_{M_\phi}]_M} | \\
    &\le \epsilon_R + \gamma \sup_{x,a} |\ip{P_\phi(\phi(x), a), V^*_{M_\phi}} - \ip{P(x,a), [V^*_{M_\phi}]_M}| \\
    &= \epsilon_R + \gamma \sup_{x,a} |\ip{P_\phi(\phi(x), a), V^*_{M_\phi}} - \ip{\Phi P(x,a), V^*_{M_\phi}}| \\
    &\le \epsilon_R + \gamma \epsilon_P \norm{V^*_{M_\phi} - \frac{R_{\text{max}}}{2(1-\gamma)}\1}_\infty \\
    &\le \epsilon_R + \gamma \epsilon_P R_{\text{max}}/(2(1-\gamma)) \\
    &= \epsilon_{R} + \gamma \Big({\displaystyle \sum^{d}_{i=1} \epsilon_{i, P}}\Big) R_{\text{max}} / (2(1 -\gamma))
\end{align*}
\endgroup

\clearpage
\section{Implementation Details}

\subsection{Linear MDP}
\label{subsec: linear-mdp}
We consider a discrete MDP with three state variables $x^1, x^2, x^3$, all taking values in $[-10, 10]$ and a single action $a$. The transition probabilities of each $x^i_t$ depend only on $x_{t-1}^i$ and the action $a$. We define the three policies/environments as ones that lead to soft interventions (interventions on the noise variables) on the three state variables individually. Each episode ends after $10$ time steps so as to have data from multiple trajectories, as is the case for more complex MDPs. We note that around $1000$ samples of data from each environment is enough for ICP to extract the correct parent sets for each state variable. Next, we estimate the transition probabilities for a particular transition using the invariant as well as the MLE based estimators and plot them against the number of samples used.

\subsection{PyTorch-like Pseudocode for Learning Model-Invariant Representations}
\label{subsec: pseudocode}
\begin{lstlisting}
for x in loader: # load a minibatch x with n samples
    # independent predictions from two randomly initiated models
    z1, z2 = f(x), h(x) # f: model_1, h: model_2
    # pick random dimension
    dim = rand(z1.shape) 
    
    pred_1 = g(cat(z1[dim], one_hot(dim))) # g: critic
    pred_2 = g(cat(z2[dim], one_hot(dim))) 
    p1, p2 = InvLoss(pred_1, pred_2)
    
    L = KL(p1, p2)
    
    L.backward()
    update(f, h, g)
    
def InvLoss(pred_1, pred_2):
    phi_1 = pred_1 * pred_2.T.detach()
    phi_2 = pred_2 * pred_1.T.detach()
    
    # matrix of inner product of 2-norm of pred_1 rows with pred_2 columns
    norm_12 = normalize(pred_1, pred_2) 
    phi_1 = phi_1 / norm_12
    phi_2 = phi_2 / norm_12.T
    
    p1 = F.softmax(phi_1, dim=-1) 
    p2 = F.softmax(phi_2, dim=-1)
    return p1, p2
    
def KL(p1, p2):
    p2 = p2.detach()
    return (p1 * (p1 / p2).log()).sum(dim=-1).mean()
\end{lstlisting}

\subsection{SAC-SVG Algorithm}
\label{appsec: sac-svg}
The SAC-SVG algorithm is presented in~\citet{amos2020model} and is based on the idea of model-based value expansion (MVE)~\citep{feinberg2018model}. MVE uses the model to expand the value function to compute a multi-step estimate which a model-free base algorithm uses for policy optimization. In SAC-SVG, the model-free base learner is a SAC agent and the multi-step estimates correspond to that of the $Q$ value used by the SAC actor. 

\begin{align*}
    \mathcal{L}^{\text{SAC-SVG}}_{\alpha, \pi} = \mathbb{E}_{x \sim \mathcal{D}, \ a \sim \pi} - Q^{\alpha, \pi}_{0:H}(x, a),
\end{align*}

where $\alpha$ is the entropy temperature parameter of SAC. Note that for $H=0$, SAC-SVG is equivalent to SAC, since the model is no longer used for updating the actor. Thus the impact of the model on the final algorithm performance is through the horizon parameter $H$. Regarding the model learner, SAC-SVG uses a recurrent deterministic model which takes as input the current state and a hidden state to output the next state for a given horizon step $H$. The other popular alternative is to use an ensemble of probabilistic model learners, as done in~\citet{chua2018deep}.

\subsection{MBPO vs POPLIN Environments}
\label{appsec: mbpo-envs}
For our MBRL experiments, we used two sets of MuJoCo-based environments, each used before in individual papers. Specifically, the POPLIN based environments were originally used in the paper by ~\citep{wang2019exploring}. These refer to the `-v0' versions from OpenAI Gym~\citep{brockman2016openai} and also includes a separately tweaked Cheetah (called PETS-Cheetah) and Swimmer environments. On the other hand, the MBPO based environments refer to the ones used by the paper ~\citep{janner2019trust} and largely correspond to the `-v2' versions from OpenAI Gym. These include an additional reward for staying alive throughout an episode.

\subsection{Spurious Correlation}
\label{appsec: spurious-correlation}
For the experiment in Section~\ref{subsec: spurious-correlation}, we used three different input strategies to test for the presence of spurious correlations in model learning. Here, we define the exact masking schemes used. We are interested in only predicting a single dimension here--- the left knee joint position. Below are the masking detailed descriptions:

\begin{itemize}
    \item \textbf{No Mask:} None of the observation dimensions are masked.
    \item \textbf{Mask:} Dimensions that are seemingly uncorrelated to the left knee joint are masked. Specifically, \{left\_shoulder\_1, left\_shoulder\_2, left\_elbow\} (qpos and qvel)
    \item \textbf{Mask\_2:} Dimensions that are seemingly correlated to the left knee joint are masked. Specifically, \{left\_hip\_x, left\_hip\_y, left\_hip\_z, left\_knee\} (qpos and qvel)
\end{itemize}

\begin{table}[t]
\begin{flushleft}
\resizebox{\linewidth}{!}{%
\begin{tabularx}{\textwidth} { >{\raggedright\arraybackslash}ll } \toprule
    \textbf{Hyperparameter} & \textbf{Value} \\ 
    \bottomrule
    
    Replay buffer size & $1000000$ \\
    
    Initial temperature ($\alpha$) & $0.1$ \\
   
    Learning rate & $1e-4$ SAC actor and critic; $1e-3$ Model learner  \\
    SAC Critic $\tau$ & 0.005 \\
    Discount $\gamma$ & $0.99$  \\
    SAC batch size & $1024$ \\
    Model batch size & $512$ \\
    Optimizer & Adam \\
    Model updates per env step & $4$ \\
    Initial steps & $1000$ \\
    Number of encoder hidden layers (Model) & $2$ \\
    Number of decoder hidden layers (Model) & $2$ \\
    Encoder hidden layer size (Model) & $512$ \\
    Decoder hidden layer size (Model) & $512$ \\
    Model critic ($g$) & Single layer MLP ($512$) \\
    \bottomrule
\end{tabularx}}
\end{flushleft}
\caption{Hyper-parameters used for the Invariant-SAC-SVG algorithm.}
\label{tab:Hyperparams-SAC}
\end{table}

\subsection{Invariant Model Learning}

For our invariant model learner, we test on offline data collected in a replay buffer during the first 1M training steps of a model-free SAC agent. We start model training with the initial samples from the replay buffer and continue to add more as the training progresses. Such a scheme ensures that we have access to changing state distributions as the policy changes while remaining isolated from direct policy optimization on the CDP. 

\clearpage
\subsection{Invariant MBRL Architecture}

\begin{figure}[t]
    \centering
    \includegraphics[angle=-90, scale=0.42]{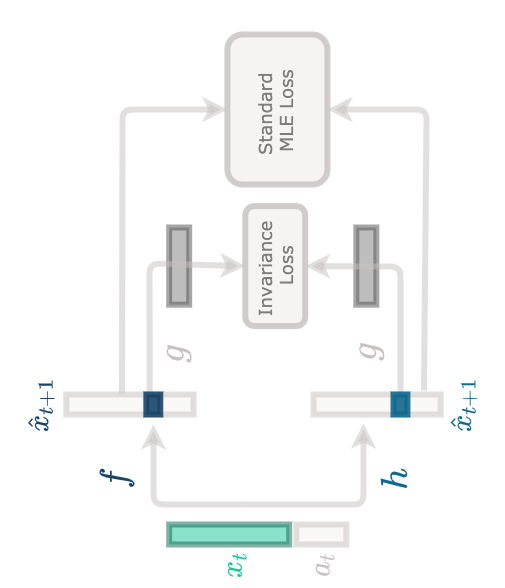}
    \caption{Architecture for learning model-invariant representations.}
    \label{fig: deep-model}
\end{figure}

Figure~\ref{fig: deep-model} uses two transition dynamics models, denoted by $f$ and $h$. Both are initialized independently and trained on independent data. $g$ refers to the critic function, which provides a score for a chosen dimension of the output of $f$ and $h$ models. This constitutes the invariance loss. The output of each transition model is also used for the standard MLE loss, which compares the predicted next state and the observed next state.

\section{Ablations}

\subsection{Per-state vs Per-state-variable Invariance}

\begin{figure}[ht]
    \centering
    \includegraphics[width=0.3\linewidth]{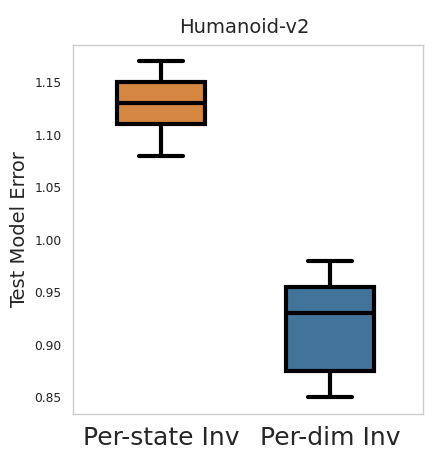}
    \vspace{5pt}
    \caption{Test modelling error when we enforce per-state vs per-dim invariance on the Humanoid-v2 task.}
    \label{fig: rebuttal}
\end{figure}

As noted in Section~\ref{sec: practical-model-invariance}, we enforce the invariance objective on a per-state-variable (per-dim) level. The rationale for this is that we wish to find the invariant predictions for the transition dynamics of each state variable. Here, we compare this approach with when we enforce invariance at a per-state level instead. 

\end{document}